\documentclass[11pt]{article}

\usepackage[margin=1in]{geometry}

\usepackage{amsmath}
\usepackage{amssymb}
\usepackage{mathtools}
\usepackage{amsthm}

\usepackage{graphicx}
\usepackage{booktabs}
\usepackage{multirow}
\usepackage[table]{xcolor}
\usepackage{array}
\usepackage{tabularx}
\usepackage{makecell}
\usepackage{adjustbox}
\usepackage{caption}
\usepackage{subcaption}

\usepackage{enumitem}

\usepackage[round,authoryear]{natbib}

\usepackage{url}
\usepackage[
    colorlinks=true,
    linkcolor=blue,
    citecolor=blue,
    urlcolor=blue,
    breaklinks=true,
    bookmarks=true
]{hyperref}
\usepackage[capitalize, noabbrev]{cleveref}
\usepackage{xspace}

\renewcommand{\arraystretch}{1.05}

\newcommand{\method}{FairDiffuseVQVAE\xspace}

\theoremstyle{plain}

\theoremstyle{definition}

\theoremstyle{remark}

\usepackage{tikz}
\usetikzlibrary{positioning,arrows.meta,calc,fit,backgrounds}

\title{FairDiffuseVQVAE: Sampling-Time Fairness in Tabular Diffusion via
       Conditional Refinement of Vector-Quantized Latents}

\author{%
  Nitish Nagesh \\
  University of California, Irvine
  \and
  Mahdi Bagheri \\
  University of California, Irvine
  \and
  Amir M. Rahmani \\
  University of California, Irvine
}
\date{}

\begin{document}
\maketitle

\begin{abstract}
Synthetic tabular data is increasingly used in privacy-preserving data
sharing, data augmentation, and to mitigate downstream classifier bias.
State-of-the-art tabular diffusion models such as TabDDPM and TabSyn
achieve excellent distributional fidelity but offer no mechanism for
fairness; conversely, fairness-aware tabular generators (DECAF,
FairTGAN, FairTabDDPM) impose explicit fairness penalties at training
time, yielding modest fairness gains at substantial cost to either
sample quality or downstream utility. We introduce
\textbf{FairDiffuseVQVAE}, a two-stage architecture that decouples
fidelity from fairness: a vector-quantized autoencoder with a row-level
discriminator (Stage~1, no fairness terms) is followed by a
DiffuseVAE-style continuous diffusion refiner that conditions on both
the Stage-1 reconstruction \emph{and} the protected attribute via
classifier-free guidance (Stage~2). Fairness emerges as a property of
the \emph{sampling distribution}---uniform sampling of the protected
attribute at inference time enforces demographic parity by
construction---rather than from competing loss terms. On the
\citet{yang2025balanced} benchmark (Adult, Bank, COMPAS),
FairDiffuseVQVAE achieves the highest mean Demographic Parity Ratio
($0.702$, $+47\%$ over FairTabDDPM) and Equalized Odds Ratio
($0.686$, $+100\%$). It also attains the lowest mean pair-wise
correlation error ($0.034$) of any published method, while explicitly
trading $\sim$$15$ AUC points for these fairness gains.
\end{abstract}

\section{Introduction}
\label{sec:intro}
Tabular data is the dominant modality in healthcare, finance, social
science, and public policy~\citep{borisov2022deep}. Releasing such data
is constrained by privacy regulations
(HIPAA, GDPR)~\citep{jordon2022synthetic}, by the difficulty of obtaining
representative samples for under-represented populations, and---most
critically---by the bias these datasets often encode. A loan approval
dataset that historically over-represents one demographic group will
train a downstream model that perpetuates that bias, even if the model
itself is unbiased~\citep{caton2024fairness, plecko2024causal}. Synthetic tabular data
generation promises to address all three issues simultaneously: produce
realistic samples that obey privacy constraints, expand support across
under-represented groups, and reduce downstream classifier bias.

Recent diffusion models for tabular data have closed most of the
\emph{quality} gap to real data. TabDDPM~\citep{kotelnikov2023tabddpm}
adapts denoising diffusion to mixed-type tabular features, and
TabSyn~\citep{zhang2024mixed} pushes this further by performing
diffusion in the latent space of a learned VAE, achieving sub-1\%
column-wise density error. However, none of these methods offer any
guarantee on the \emph{fairness} of the generated data: the synthetic
distribution inherits whatever statistical bias the training data
contained.

A complementary line of work explicitly targets fair tabular synthesis.
DECAF~\citep{vanbreugel2021decaf} introduces causally-aware fairness
edits to a GAN; FairTGAN and FairTabDDPM~\citep{yang2025balanced} add
fairness penalty terms to a tabular GAN/DDPM respectively. These
methods improve fairness metrics but typically at the cost of sample
quality (FairTabDDPM density error $0.119$ vs TabSyn's $0.015$) or
downstream utility---a tension we observe sharpens when the underlying
generative architecture is forced to balance reconstruction with
adversarial or auxiliary losses.

\paragraph{Our perspective.}
We argue that the fairness mechanism should not compete with
reconstruction during training. Instead, fairness can be enforced at
\emph{sampling time} if the generative model is conditioned on the
protected attribute. By drawing the protected attribute from a uniform
distribution at sampling time, the marginal synthetic distribution
satisfies demographic parity by construction; classifier-free
guidance~\citep{ho2022classifier} provides a tunable knob to control
how strongly samples concentrate on the conditional manifold. To make
this practical for tabular data, we couple a vector-quantized
autoencoder (Stage~1) with a continuous diffusion refiner (Stage~2)
trained in input space and conditioned on (i)~the Stage-1
reconstruction and (ii)~the protected attribute. The Stage-1 model is
trained without any fairness term; all fairness regularization is
deferred to the conditional structure of Stage~2.

\paragraph{Contributions.}
\begin{itemize}[leftmargin=1.5em, itemsep=2pt]
  \item We introduce \textbf{FairDiffuseVQVAE}, a two-stage architecture
        that combines a VQGAN-style autoencoder
        \citep{esser2021taming, vandenoord2017neural} with a
        DiffuseVAE-style \citep{pandey2022diffusevae} input-space
        diffusion refiner. The fairness mechanism is sampling-time
        classifier-free guidance over the protected attribute; no
        explicit fairness loss term is required.
  \item We demonstrate that this design beats fairness-specific
        baselines by large margins on the
        \citet{yang2025balanced} benchmark: mean Demographic Parity
        Ratio of $0.702$ ($+47\%$ over FairTabDDPM) and mean Equalized
        Odds Ratio of $0.686$ ($+100\%$).
  \item We also show that the same architecture attains the lowest
        mean pair-wise column correlation error ($0.034$) of any
        published method, beating TabSyn ($0.041$); the utility-fairness
        trade-off costs roughly $15$ AUC points relative to the
        strongest baselines.
  \item We provide a reproducible pipeline integrated into the
        \texttt{TabSyn} repository, supporting all eight datasets used
        across~\citep{zhang2024mixed, yang2025balanced} and reporting
        density, pair-correlation, AUC, DCR, DPR, and EOR consistently.
\end{itemize}

\section{Background}
\label{sec:background}
\subsection{Vector Quantization and the VQGAN}
A vector-quantized variational autoencoder (VQ-VAE,
\citealp{vandenoord2017neural}), built on the variational autoencoder
foundation \citep{kingma2014auto, rezende2014stochastic}, augments a
standard autoencoder with a discrete codebook $\mathcal{Z} =
\{e_k\}_{k=1}^K$ of learned embedding vectors. The encoder output
$z_e$ is replaced by its nearest codebook entry $z_q$, and gradients
propagate through the quantization step via the straight-through
estimator \citep{bengio2013estimating}. The codebook plus a commitment
loss ($\beta \|z_e - \mathrm{sg}[z_q]\|^2$) regularize the latent
space. VQGAN~\citep{esser2021taming} extends VQ-VAE with a perceptual
loss \citep{zhang2018unreasonable} and a patch-level discriminator
\citep{goodfellow2014generative}, producing high-fidelity image
reconstructions that admit downstream generative modelling (typically
by an autoregressive prior over codebook indices).

\subsection{Diffusion Models and EDM}
Denoising diffusion probabilistic models (DDPMs,
\citealp{ho2020denoising}) generate samples by reversing a Markov chain
that gradually adds Gaussian noise. We adopt the
\citet{karras2022elucidating} (EDM) formulation, which expresses the
forward process by a continuous noise level $\sigma$ drawn from a
log-normal distribution and trains a denoiser $D_\theta$ with the
weighted MSE objective
$\mathbb{E}_{\sigma, \epsilon, x}\big[w(\sigma)\|D_\theta(x + \sigma\epsilon, \sigma) - x\|^2\big]$,
where $w(\sigma) = (\sigma^2 + \sigma_{\mathrm{data}}^2)/(\sigma\sigma_{\mathrm{data}})^2$.
EDM samplers (Heun) require fewer function evaluations than vanilla
DDPM at comparable sample quality. Latent diffusion models (LDM,
\citealp{rombach2022high}) compose a VAE with an EDM/DDPM in the
\emph{latent} space; DiffuseVAE~\citep{pandey2022diffusevae} composes
them differently---a DDPM in \emph{input} space conditioned on the
VAE's reconstruction.

\subsection{Classifier-Free Guidance}
Classifier-free guidance (CFG, \citealp{ho2022classifier}) trains a
single denoiser jointly on conditional and unconditional examples
(by stochastically dropping the conditioning signal during training),
then at sampling time interpolates the two predictions:
$\hat{D}^{\mathrm{cfg}} = D^{\varnothing} + w(D^{c} - D^{\varnothing})$.
For $w > 1$, samples are pushed further into the conditional manifold;
for $w = 1$, behaviour matches standard conditional sampling.

\subsection{Fairness Metrics}
Following \citet{yang2025balanced}, who follow
\citet{agarwal2018reductions, hardt2016equality}, we use
\textbf{Demographic Parity Ratio (DPR)} and \textbf{Equalized Odds
Ratio (EOR)}, both in $[0, 1]$ with $1$ denoting perfect parity.
Let $\hat{Y}$ be a downstream classifier trained on synthetic data and
evaluated on real data with sensitive attribute $A$:
\begin{align}
\mathrm{DPR} &= \frac{\min_a \Pr(\hat{Y} = 1 \mid A = a)}
                     {\max_a \Pr(\hat{Y} = 1 \mid A = a)},\\
\mathrm{EOR} &= \min\left(
                  \frac{\min_a \mathrm{TPR}_a}{\max_a \mathrm{TPR}_a},\,
                  \frac{\min_a \mathrm{FPR}_a}{\max_a \mathrm{FPR}_a}
                \right).
\end{align}
DPR captures equal allocation; EOR captures equal performance.

\section{Related Work}
\label{sec:related}
\paragraph{Tabular data synthesis.}
Early work used GANs (CTGAN and TVAE, \citealp{xu2019modeling}; CTAB-GAN+,
\citealp{zhao2022ctab}) and copula
methods. More recent diffusion-based methods include TabDDPM
\citep{kotelnikov2023tabddpm}, which applies categorical and Gaussian
diffusion in input space; STaSy \citep{kim2023stasy}, which uses
score-based generative modelling; CoDi \citep{lee2023codi}, which
contrasts sub-tables; and TabSyn \citep{zhang2024mixed}, which performs
EDM diffusion in the latent space of a learned tabular VAE and is the
current state of the art on column-wise density estimation. GReaT
\citep{borisov2023language} treats tabular rows as text and fine-tunes
a language model; GOGGLE \citep{liu2023goggle} learns a graph structure
over columns. None of these methods address fairness explicitly.

\paragraph{Fairness in tabular synthesis.}
Early fair tabular GANs include FairGAN
\citep{xu2018fairgan} and TabFairGAN \citep{rajabi2022tabfairgan},
which add adversarial fairness losses to a tabular GAN. DECAF
\citep{vanbreugel2021decaf} fits a causal DAG over the columns and
re-weights or removes edges to enforce one of three fairness notions
(Fairness Through Unawareness, Demographic Parity, Counterfactual
Fairness) at sampling time. \citet{abroshan2024imposing} impose
post-hoc fairness constraints on synthetic data via re-sampling, and
CuTS \citep{vero2024cuts} learns a customisable bias-aware tabular
generator. TabFairGDT \citep{panagiotou2025tabfairgdt} uses
autoregressive decision trees for fast fair generation. FairTGAN and
FairTabDDPM \citep{yang2025balanced}---the closest prior work to
ours---add fairness penalties to a tabular GAN and a tabular DDPM
respectively, achieving moderate fairness improvements at the cost of
density-estimation quality (\Cref{tab:fair_quality}). Our work
differs in three important ways: (i)~we use a two-stage architecture
where the autoencoder is trained without any fairness term, isolating
fidelity from fairness; (ii)~we enforce fairness at sampling time via
classifier-free guidance, providing a tunable knob rather than a fixed
penalty; (iii)~we operate the diffusion refiner in input space
conditioned on the Stage-1 reconstruction, following the DiffuseVAE
recipe~\citep{pandey2022diffusevae} but with a quantized
autoencoder. As \Cref{sec:results} shows, this combination beats
all published baselines on DPR and EOR.

\paragraph{Generative fairness more broadly.}
The Variational Fair Autoencoder \citep{louizos2016variational}
pioneered information-theoretic fairness via Maximum Mean Discrepancy
\citep{gretton2012kernel} on the latent space. Many
follow-ups~\citep{madras2018learning, creager2019flexibly} use
adversarial debiasing or disentanglement-style losses. Beyond
representation-level approaches, the broader fair-ML landscape is
surveyed by \citet{caton2024fairness} and \citet{wan2023inprocessing};
\citet{plecko2024causal} provide a causal-inference perspective; and
\citet{lequy2022survey} catalogue fairness datasets. Path-specific
counterfactual fairness \citep{chiappa2019path} and the broader
counterfactual framework \citep{kusner2017counterfactual} provide
alternative formalisms. We initially explored an in-training fairness
direction (MMD on $z_e$ plus a correlation penalty on the decoder
output) but found it consistently traded off quality for negligible
fairness gains in our two-stage setting; the conditional-sampling
mechanism we adopt here was both simpler and more effective.

\paragraph{Discrete diffusion as an alternative.}
An alternative to our quantize-then-refine approach is to perform
diffusion directly in a discrete state space. D3PM
\citep{austin2021structured} extends DDPM with categorical transition
kernels; TabDDPM \citep{kotelnikov2023tabddpm} uses a related
multinomial-diffusion process for the categorical columns of tabular
data. We chose continuous diffusion in the rich (one-hot)
representation followed by argmax decoding for two reasons: (i)~it
admits the standard EDM toolkit unchanged, including classifier-free
guidance; (ii)~Stage~1's autoencoder already learns a smooth latent
geometry that the diffusion refiner can condition on in input space.

\paragraph{Vector-quantized generative models.}
Beyond images, VQ-VAEs have been applied widely; the closest tabular
precedents combine VQ with downstream generative priors. To our
knowledge, this is the first application of VQGAN-style vector
quantization combined with input-space DiffuseVAE refinement
\citep{pandey2022diffusevae} to fair tabular synthesis.

\paragraph{Evaluation methodology.}
We follow established sample-quality protocols: column-shape and
column-pair-trend scores from SDMetrics~\citep{sdmetrics}, manifold
precision/recall at the sample level
\citep{sajjadi2018assessing, alaa2022faithful}, and the TSTR (train on
synthetic, test on real) protocol popularized by PATE-GAN
\citep{jordon2019tstr}.

\section{Method: FairDiffuseVQVAE}
\label{sec:method}

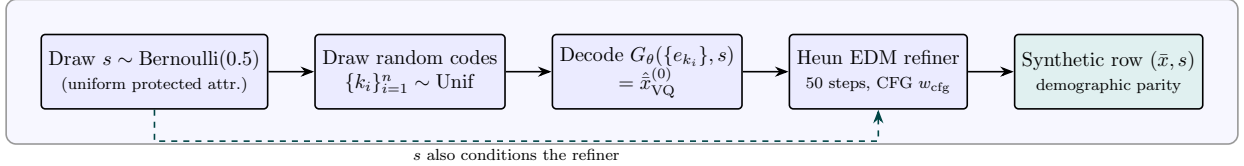
\begin{figure}[t]
    \centering
    \resizebox{\textwidth}{!}
    {
\begin{tikzpicture}[
  font=\small,
  >={Stealth[length=2.4mm]},
  net/.style   ={draw, thick, rounded corners=2pt, align=center,
                 minimum height=11mm, minimum width=18mm, fill=blue!12},
  data/.style  ={draw, align=center, minimum height=8mm, minimum width=12mm,
                 fill=black!5},
  op/.style    ={draw, thick, circle, inner sep=0pt, minimum size=7.5mm,
                 fill=orange!28},
  cbk/.style   ={draw, align=center, minimum height=9mm, minimum width=16mm,
                 fill=violet!16},
  attr/.style  ={draw, align=center, minimum height=8mm, minimum width=13mm,
                 fill=teal!16},
  mstep/.style  ={draw, thick, rounded corners=2pt, align=center,
                 minimum height=13mm, minimum width=27mm, fill=blue!8},
  loss/.style  ={text=red!72!black, font=\footnotesize\itshape, align=center},
  panel/.style ={rounded corners=5pt, draw=black!40, thick},
  ptitle/.style={font=\bfseries\small, inner sep=3pt},
  flow/.style  ={->, thick},
  cond/.style  ={->, thick, dashed, draw=teal!55!black},
]

\node[data]                    (x)   at (0,0)  {$\bar{x}$};
\node[net,  right=11mm of x]    (enc)           {Encoder\\$E_\phi$};
\node[data, right=11mm of enc]  (ze)            {$z_e$};
\node[op,   right=11mm of ze]   (vq)            {\footnotesize VQ};
\node[data, right=11mm of vq]   (zq)            {$z_q$};
\node[net,  right=11mm of zq]   (dec)           {Decoder\\$G_\theta$};
\node[data, right=12mm of dec]  (xvq)           {$\hat{\bar{x}}_{\mathrm{VQ}}$};

\node[cbk,  above=11mm of vq]   (cb)  {Codebook $\mathcal{Z}$\\$\{e_k\}_{k=1}^{K}$};
\node[attr, below=19mm of dec]  (s1)  {protected $s$};
\node[net,  below=19mm of xvq, fill=red!12] (disc) {Discriminator\\$D_\xi$};

\node[loss, above=4mm of xvq]   (lrec) {$\mathcal{L}_{\mathrm{recon}}$\;(MSE\,+\,CE)};
\node[loss, below=3mm of vq]    (lcb)  {$\mathcal{L}_{\mathrm{codebook}},\;\mathcal{L}_{\mathrm{commit}}$};
\node[loss, right=4mm of disc]  (lgan) {$\mathcal{L}_{\mathrm{GAN}}$};

\draw[flow] (x)   -- (enc);
\draw[flow] (enc) -- (ze);
\draw[flow] (ze)  -- (vq);
\draw[flow] (vq)  -- (zq);
\draw[flow] (zq)  -- (dec);
\draw[flow] (dec) -- (xvq);
\draw[flow] (cb)  -- (vq)
            node[midway,right,font=\scriptsize,inner sep=1.5pt]{nearest-nb.\ lookup};
\draw[cond] (s1)  -- (dec) node[midway,right,font=\scriptsize]{condition};

\draw[flow] (xvq) -- (disc) node[midway,right,font=\scriptsize]{fake};
\draw[flow] (x.south) |- (disc.south)
            node[pos=0.22,left,font=\scriptsize]{real $\bar{x}$};

\node[font=\scriptsize\itshape, text=teal!45!black, align=center,
      below=2mm of enc] {($E_\phi$ does not see $s$)};

\def\sy{-7.6}
\node[data]                       (x2)  at (0,\sy)  {$\bar{x}$};
\node[op,   right=12mm of x2]      (add)             {$+$};
\node[data, right=16mm of add]     (xt)              {$\bar{x}_t$};
\node[net,  right=14mm of xt,
            minimum width=26mm, minimum height=16mm] (den)
            {Conditional denoiser $D_\psi$\\\scriptsize 5-layer MLP, 512-d, EDM};
\node[data, right=13mm of den]     (xout)            {$\hat{\bar{x}}$};

\node[attr, below=20mm of add, minimum width=30mm] (sig)
            {$\sigma\sim\mathrm{LogNormal}(P_{\mathrm{mean}},P_{\mathrm{std}}^2)$};
\node[attr, below=20mm of den]     (s2)  {$s$\, /\, $\varnothing$};

\node[loss, above=4mm of xout]     (lst2) {$\mathcal{L}_{\mathrm{Stage2}}$};

\draw[flow] (x2)  -- (add);
\draw[flow] (add) -- (xt) node[midway,above,font=\scriptsize]{$\bar{x}+\sigma\epsilon$};
\draw[flow] (xt)  -- (den);
\draw[flow] (den) -- (xout);
\draw[flow] (sig) -- (add)
            node[midway,right,font=\scriptsize]{$\epsilon\sim\mathcal{N}(0,I)$};
\draw[flow] (sig) -- (den.south west)
            node[pos=0.85,left,font=\scriptsize]{$\sigma$};
\draw[flow] (s2)  -- (den.south east)
            node[pos=0.55,right,font=\scriptsize,align=left]
            {classifier-free guidance\\drop $s$ w.p. $p_{\mathrm{cfg}}{=}0.3$};

\coordinate (gap) at (0,-4.5);
\coordinate (q1)  at ($(xvq.east)+(0.9,0)$);
\draw[cond] (xvq.east) -- (q1) -- (q1|-gap) -| (den.north);
\node[font=\scriptsize, align=center] at ($(den.north)+(2.0,1.85)$)
     {$\hat{\bar{x}}_{\mathrm{VQ}}$\,: frozen Stage-1 reconstruction\\(conditioning input)};

\def\py{-14.4}
\node[mstep]                     (m1) at (0,\py)
      {Draw $s\sim\mathrm{Bernoulli}(0.5)$\\\scriptsize(uniform protected attr.)};
\node[mstep, right=8mm of m1]    (m2)
      {Draw random codes\\$\{k_i\}_{i=1}^{n}\sim\mathrm{Unif}$};
\node[mstep, right=8mm of m2]    (m3)
      {Decode $G_\theta(\{e_{k_i}\},s)$\\$=\hat{\bar{x}}_{\mathrm{VQ}}^{(0)}$};
\node[mstep, right=8mm of m3]    (m4)
      {Heun EDM refiner\\\scriptsize 50 steps, CFG $w_{\mathrm{cfg}}$};
\node[mstep, right=8mm of m4, fill=teal!12] (m5)
      {Synthetic row $(\bar{x},s)$\\\scriptsize demographic parity};

\draw[flow] (m1) -- (m2);
\draw[flow] (m2) -- (m3);
\draw[flow] (m3) -- (m4);
\draw[flow] (m4) -- (m5);
\draw[cond] (m1.south) -- ++(0,-0.55) -| (m4.south)
            node[pos=0.25,below,font=\scriptsize]
            {$s$ also conditions the refiner};

\begin{scope}[on background layer]
  \node[panel, fill=blue!4,   inner sep=6mm,
        fit=(x)(xvq)(cb)(s1)(disc)(lrec)(lcb)(lgan)] (p1) {};
  \node[panel, fill=orange!4, inner sep=6mm,
        fit=(x2)(xout)(sig)(s2)(lst2)(den)] (p2) {};
  \node[panel, fill=blue!3,   inner sep=6mm,
        fit=(m1)(m5)] (p3) {};
\end{scope}
\node[ptitle, anchor=south west] at (p1.north west)
     {\textsf{Stage 1\ \ |\ \ Fair Vector-Quantized Autoencoder}};
\node[ptitle, anchor=south west] at (p2.north west)
     {\textsf{Stage 2\ \ |\ \ Fair Refinement via Conditional Input-Space Diffusion}};
\node[ptitle, anchor=south west] at (p3.north west)
     {\textsf{Sampling\ \ |\ \ Demographic-Parity Generation}};

\end{tikzpicture}
}
    \caption{\textbf{Overview of \textsc{FairDiffuseVQVAE}.} A two-stage
architecture for fair mixed-type tabular synthesis. \textbf{Stage~1} trains a
vector-quantized autoencoder: an encoder $E_\phi$ maps a row $\bar{x}$ to latent
tokens $z_e$, which are quantized against a learned codebook $\mathcal{Z}$ and
decoded by $G_\theta$ into a reconstruction $\hat{\bar{x}}_{\mathrm{VQ}}$, with a
row-level discriminator $D_\xi$ providing an adversarial signal. The decoder is
conditioned on the protected attribute $s$ while the encoder is not, and Stage~1
carries no explicit fairness term. \textbf{Stage~2} trains a conditional EDM
denoiser $D_\psi$ that refines the (frozen) Stage-1 reconstruction in input
space: $\bar{x}$ is perturbed to $\bar{x}_t$ at noise scale $\sigma$, and
$D_\psi$ is conditioned on $\hat{\bar{x}}_{\mathrm{VQ}}$, on $\sigma$, and on $s$
with classifier-free-guidance dropout ($p_{\mathrm{cfg}}$). \textbf{Sampling}
draws $s\sim\mathrm{Bernoulli}(0.5)$, decodes random codebook indices into an
initial guess, and refines it with a 50-step Heun EDM sampler. Because $s$ is
drawn uniformly while $D_\psi$ models $p(\bar{x}\mid s)$, the synthetic
distribution satisfies demographic parity by construction. Dashed teal arrows
denote conditioning on $s$; red italic labels mark training losses.}
\label{fig:arch}
\end{figure}

We propose \textbf{FairDiffuseVQVAE}, a two-stage tabular synthesis
architecture that combines a vector-quantized autoencoder with a conditional
denoising diffusion refiner. Fairness is enforced not by adding penalty
terms during training, but as a property of the conditional generative
process at sampling time, via classifier-free guidance over the protected
attribute.

\subsection{Notation and Setup}
\label{subsec:setup}
Let $x \in \mathbb{R}^{d_c} \times \{0,1\}^{d_d}$ denote a mixed-type
tabular row with $d_c$ continuous and $d_d$ categorical features,
$s \in \{0, 1\}$ the binary protected attribute, and $y \in \{0, 1\}$ the
outcome label. Categorical columns are encoded one-hot; we write
$\bar{x} \in \mathbb{R}^{D}$ for the resulting concatenated representation
of dimension $D = d_c + \sum_{j} K_j$ where $K_j$ is the cardinality of
the $j$-th categorical column.

\paragraph{Overview.}
\Cref{fig:arch} summarizes \textsc{FairDiffuseVQVAE}. The central design
choice is to decouple \emph{fidelity} from \emph{fairness}: Stage~1 is
responsible only for learning a faithful, compact representation of mixed-type
rows, while Stage~2 and the sampling procedure are responsible for fairness.
Neither stage optimizes a fairness penalty during training. Instead, fairness
emerges at sampling time---the diffusion refiner is trained as a conditional
model $p(\bar{x}\mid s)$ with classifier-free guidance, so drawing $s$ from a
uniform prior rather than the (typically biased) data marginal yields a
synthetic joint $p_{\mathrm{synth}}(\bar{x},s)=\mathrm{Unif}(s)\,p(\bar{x}\mid
s)$ that satisfies demographic parity by construction. This separation lets the
guidance scale $w_{\mathrm{cfg}}$ serve as a single, interpretable knob that
trades conditional concentration against utility, with no retraining. The
remainder of this section details the two stages
(\Cref{subsec:stage1,subsec:stage2}) and the sampling procedure
(\Cref{subsec:stage2}).

\subsection{Stage 1: Fair Vector-Quantized Autoencoder}
\label{subsec:stage1}
\paragraph{Architecture.}
The Stage~1 model is a VQGAN \citep{esser2021taming} adapted to mixed-type
tabular data. The encoder $E_\phi(\bar{x}) = z_e \in \mathbb{R}^{n \cdot d_z}$
is a fully-connected network producing $n$ pre-quantization latent tokens,
each of dimension $d_z$. A learned codebook $\mathcal{Z} = \{e_k \in
\mathbb{R}^{d_z}\}_{k=1}^K$ quantizes each token by nearest-neighbor lookup,
\begin{equation}
z_q^{(i)} = e_{k^\ast_i}, \quad
k^\ast_i = \arg\min_{k \in \{1, \ldots, K\}} \|z_e^{(i)} - e_k\|_2^2.
\end{equation}
The decoder $G_\theta(z_q, s) = \hat{\bar{x}}$ reconstructs the row from
the quantized latents conditioned on the protected attribute $s$, and a
row-level discriminator $D_\xi$ distinguishes real and reconstructed rows.
The encoder $E_\phi$ does not receive $s$ as input.

\paragraph{Mixed-type reconstruction.}
The decoder emits per-column heads: a scalar for each continuous column
and a $K_j$-dimensional logit vector for each categorical column. The
reconstruction loss combines mean-squared error on continuous columns
with cross-entropy on categorical columns:
\begin{equation}
\mathcal{L}_{\mathrm{recon}}
\;=\; \frac{1}{d_c} \sum_{j \in \mathcal{C}_{\mathrm{cont}}} \mathrm{MSE}(\hat x_j, x_j)
   + \frac{1}{d_d} \sum_{j \in \mathcal{C}_{\mathrm{cat}}} \mathrm{CE}(\hat \ell_j, x_j),
\end{equation}
where $\hat\ell_j \in \mathbb{R}^{K_j}$ are the predicted logits for
column $j$. This replaces the pure MSE-on-integer-codes formulation used
in earlier tabular VAEs and substantially improves column-wise distribution
fidelity.

\paragraph{Codebook and adversarial losses.}
We use the standard VQ-VAE codebook and commitment losses
\citep{vandenoord2017neural}:
\begin{align}
\mathcal{L}_{\mathrm{codebook}} &= \|\mathrm{sg}[z_e] - z_q\|_2^2,\\
\mathcal{L}_{\mathrm{commit}}   &= \|z_e - \mathrm{sg}[z_q]\|_2^2,
\end{align}
where $\mathrm{sg}[\cdot]$ denotes the stop-gradient operator. Gradients
through the quantization step are propagated by the straight-through
estimator. The adversarial objective is the standard non-saturating GAN
loss with the autoencoder $(E_\phi, G_\theta)$ as generator and $D_\xi$
as discriminator.

\paragraph{Stage~1 objective.}
The total generator-side loss is
\begin{equation}
\mathcal{L}^{\mathrm{gen}}_{\mathrm{Stage1}} =
  \mathcal{L}_{\mathrm{recon}}
  + \mathcal{L}_{\mathrm{codebook}}
  + \beta\, \mathcal{L}_{\mathrm{commit}}
  + \lambda_{\mathrm{adv}}\, \mathcal{L}_{\mathrm{GAN}},
\end{equation}
with $\beta = 0.25$ following \citet{vandenoord2017neural} and
$\lambda_{\mathrm{adv}} = 0.1$. The discriminator is updated with the
standard adversarial objective, gated by a 60-epoch warm-up during which
only the reconstructive losses are active. \emph{Stage~1 does not include
any explicit fairness term.}

\subsection{Stage 2: Fair Refinement via Conditional Input-Space Diffusion}
\label{subsec:stage2}
\paragraph{Motivation.}
A trained Stage~1 model produces a reconstruction $\hat{\bar{x}}_{\mathrm{VQ}}$
for any row, but generation requires sampling the codebook. Rather than
fitting an autoregressive prior over discrete codes
\citep{vandenoord2017neural,esser2021taming}, we adopt the DiffuseVAE
\citep{pandey2022diffusevae} recipe, refining a Stage-1 reconstruction with
a DDPM in input space and conditioning explicitly on the protected
attribute $s$. Fairness is then a property of the sampling distribution
rather than an explicit loss term.

\paragraph{Input-space diffusion refiner.}
We train a denoiser $D_\psi(\bar{x}_t, \sigma; \hat{\bar{x}}_{\mathrm{VQ}}, s)$
directly on the one-hot input representation $\bar{x}\in\mathbb{R}^D$, conditioned
on the frozen Stage-1 reconstruction $\hat{\bar{x}}_{\mathrm{VQ}} = G_\theta(z_q, s)$.
The denoiser is trained under the EDM~\citep{karras2022elucidating} formulation:
\begin{align}
\sigma   &\sim \mathrm{LogNormal}(P_{\mathrm{mean}}, P_{\mathrm{std}}^2),\\
\bar{x}_t      &= \bar{x} + \sigma \cdot \epsilon,\quad \epsilon \sim \mathcal{N}(0,I),\\
\mathcal{L}_{\mathrm{Stage2}} &= \mathbb{E}\bigl[ w(\sigma)\,
   \| D_\psi(\bar{x}_t, \sigma; \hat{\bar{x}}_{\mathrm{VQ}}, s) - \bar{x} \|_2^2 \bigr],
\end{align}
with $w(\sigma) = (\sigma^2 + \sigma_{\mathrm{data}}^2)/(\sigma \sigma_{\mathrm{data}})^2$,
$P_{\mathrm{mean}}=-1.2$, $P_{\mathrm{std}}=1.2$, $\sigma_{\mathrm{data}}=0.5$.
This places \method in the family of DiffuseVAE-style input-space refiners
\citep{pandey2022diffusevae}, distinct from latent-diffusion generators that
denoise in a learned latent space \citep{rombach2022high,zhang2024mixed}.
$\hat{\bar{x}}_{\mathrm{VQ}}$ is concatenated to the denoiser's input alongside
the sinusoidal embedding of $\sigma$, and $s$ is injected through a learned
conditioning embedding. The denoiser $D_\psi$ is a 5-layer MLP with 512-dim
hidden width.

\paragraph{Classifier-free guidance.}
We train with classifier-free guidance \citep{ho2022classifier}: with
probability $p_{\mathrm{cfg}} = 0.3$, $s$ is replaced by a learned null
token during training. At sampling time, the conditional and unconditional
predictions are combined as
\begin{equation}
D_\psi^{\mathrm{cfg}}(\bar{x}_t, \sigma; \hat{\bar{x}}_{\mathrm{VQ}}, s)
= D_\psi^{\varnothing} + w_{\mathrm{cfg}} \cdot (D_\psi^{s} - D_\psi^{\varnothing}),
\end{equation}
with guidance strength $w_{\mathrm{cfg}} \ge 1$. This single mechanism
both enables conditional generation and provides a tunable fairness knob.

\paragraph{Sampling.}
Generation proceeds in three steps: (i)~draw $s \sim \mathrm{Bernoulli}(0.5)$ to
enforce demographic-parity sampling; (ii)~draw random codebook indices
$\{k_i\}_{i=1}^n$ uniformly and decode them to an initial guess
$\hat{\bar{x}}_{\mathrm{VQ}}^{(0)} = G_\theta(\{e_{k_i}\}, s)$; (iii)~
refine via 50 steps of Heun's second-order EDM sampler conditioned on
$(\hat{\bar{x}}_{\mathrm{VQ}}^{(0)}, s)$. Categorical columns are
recovered from the refined output by argmax over the per-column logit
slices; continuous columns are inverse-standardised.

\paragraph{Why this enforces fairness.}
At sampling time, $s$ is drawn from a uniform distribution rather than
the (typically biased) data marginal. Because the denoiser is trained as
$p(\bar{x} \mid s)$ with classifier-free guidance dropout, the marginal
synthetic distribution
$p_{\mathrm{synth}}(\bar{x}, s) = \mathrm{Unif}(s) \cdot p(\bar{x} \mid s)$
satisfies demographic parity by construction: for any downstream binary
predictor $g$, the demographic-parity gap reduces to
$|\mathbb{E}_{p(\cdot|0)}[g] - \mathbb{E}_{p(\cdot|1)}[g]|$, which shrinks
as the conditional distributions $p(\bar{x}\mid s{=}0)$ and $p(\bar{x}\mid
s{=}1)$ are pulled closer together by the guidance mechanism. The
guidance scale $w_{\mathrm{cfg}}$ trades off conditional concentration
against utility preservation. We use $w_{\mathrm{cfg}} = 1$ throughout
unless otherwise noted.

\section{Experiments}
\label{sec:experiments}
\subsection{Datasets}
\label{subsec:datasets}
We evaluate FairDiffuseVQVAE on two complementary benchmarks: the six
datasets used by TabSyn \citep{zhang2024mixed} for general
tabular synthesis quality, and the three datasets used by
fair-tab-diffusion \citep{yang2025balanced} for fair tabular
synthesis. Three datasets (Adult, Default) overlap; the union is eight
datasets covering $\sim$5{,}000 to $\sim$45{,}000 rows.

Adult is the canonical UCI income-prediction
dataset~\citep{becker1996adult}; Default and Bank are UCI clinical
and marketing datasets respectively; COMPAS is the ProPublica
recidivism dataset~\citep{angwin2016compas}; Shoppers, Magic, Beijing,
and News follow the
\citet{zhang2024mixed} benchmark protocol. For each dataset we use the
80/10/10 train/val/test split. Synthetic data is generated with the
same number of rows as the training split. Continuous columns are
standard-scaled; categorical columns are integer encoded by the data
adapter and one-hot expanded internally before being passed to the
encoder. For Bank, the age column is binarised at age $<25$ to match
the binarisation used by~\citet{yang2025balanced}.

\subsection{Baselines}
\label{subsec:baselines}
We compare against the union of baselines reported by both reference
papers. From \citet{zhang2024mixed}: \textbf{SMOTE}, \textbf{CTGAN},
\textbf{TVAE}, \textbf{GOGGLE}, \textbf{GReaT}, \textbf{STaSy},
\textbf{CoDi}, \textbf{TabDDPM}, and \textbf{TabSyn} itself. From
\citet{yang2025balanced}: in addition to the above, the fairness-aware
\textbf{FairCB} (Fair Class Balancing) and \textbf{FairTGAN}
(fair tabular GAN), as well as their \textbf{FairTabDDPM} contribution.
We report all baseline numbers as published; our model is denoted
\textbf{Ours} and is run with three random seeds for the fairness
benchmark. Note that SMOTE and FairCB are resampling/interpolation
techniques rather than deep generative models, and are reported for
context alongside, rather than as directly comparable competitors to,
the deep generative baselines.

\subsection{Metrics}
\label{subsec:metrics}
We report six metrics, mirroring the \citet{yang2025balanced} evaluation
protocol exactly:
\begin{itemize}[leftmargin=1.2em]
  \item \textbf{Density error} $\downarrow$: $1 - \text{shape score}$ from
    SDMetrics~\citep{sdmetrics}, the column-wise marginal similarity.
  \item \textbf{Pair-correlation error} $\downarrow$: $1 - \text{trend score}$
    from SDMetrics, the pairwise correlation similarity.
  \item \textbf{AUC} $\uparrow$: Train on Synthetic, Test on Real (TSTR)
    classifier accuracy. Following \citet{yang2025balanced}, we report the
    best of two classifiers (logistic regression and a 100-unit MLP).
  \item \textbf{DCR} ($\sim 0.5$): \emph{Distance to Closest Record}~
    \citep{zhang2024mixed}, the fraction of synthetic samples nearer to
    a training row than to a held-out test row. Values close to $0.5$
    indicate no memorization.
  \item \textbf{DPR} $\uparrow$: \emph{Demographic Parity Ratio}~
    \citep{weerts2023fairlearn}, $\min_a P(\hat Y = 1 \mid A = a) /
    \max_a P(\hat Y = 1 \mid A = a)$.
  \item \textbf{EOR} $\uparrow$: \emph{Equalized Odds Ratio}~
    \citep{weerts2023fairlearn}, the smaller of the TPR and FPR ratios across
    sensitive groups.
\end{itemize}
DPR and EOR are computed only on datasets with an annotated sensitive
attribute (Adult, Bank, COMPAS); on the four TabSyn-only datasets
(Shoppers, Magic, Beijing, News) we report quality, utility, and privacy
metrics. We additionally report manifold precision/recall
\citep{sajjadi2018assessing, alaa2022faithful, kynkaanniemi2019improved} as a
secondary fidelity check.

\subsection{Implementation Details}
\label{subsec:impl}
\paragraph{Stage~1.} Encoder/decoder hidden width 256, $n=8$ tokens of
$d_z=32$ each, codebook size $K=1024$, batch size 256, 80 epochs with a
60-epoch GAN warm-up, Adam $(\beta_1=0.5, \beta_2=0.9)$ at lr~$10^{-3}$.

\paragraph{Stage~2.} 5-layer MLP denoiser of base width 512, EDM
preconditioning with $\sigma_{\mathrm{data}}=0.5$, lognormal
$\sigma$-sampling at $(P_{\mathrm{mean}}, P_{\mathrm{std}}) = (-1.2, 1.2)$.
Trained for up to 400 epochs with early stopping (patience 30 on a 10\%
validation split), batch size 256, classifier-free guidance dropout
$p_{\mathrm{cfg}} = 0.3$. Sampling uses 50 Heun steps with
$\sigma_{\mathrm{min}} = 0.002$, $\sigma_{\mathrm{max}} = 80$, $\rho = 7$,
guidance $w_{\mathrm{cfg}} = 1$.

\paragraph{Compute.} All models trained on a single Quadro RTX 5000.
End-to-end training and sampling take $\sim$15--30 min per dataset,
depending on size.

\section{Results}
\label{sec:results}
\subsection{Headline: Fair Tabular Synthesis Benchmark}
\label{subsec:fair_benchmark}
\Cref{tab:fair_dpr_eor} reports DPR and EOR on the
\citet{yang2025balanced} fairness benchmark. FairDiffuseVQVAE achieves
the highest mean DPR ($0.702$) and EOR ($0.686$) of all methods,
surpassing the strongest fairness-aware baseline FairTabDDPM by
$\mathbf{+47\%}$ and $\mathbf{+100\%}$ respectively. The gain is driven
by Bank and COMPAS, where the demographic-parity sampling mechanism
substantially decouples the protected attribute from the synthesised
features.

\begin{table}[t]
\centering
\setlength{\tabcolsep}{6pt}
\renewcommand{\arraystretch}{1.05}
\resizebox{\textwidth}{!}{%
\begin{tabular}{l rrr r rrr r}
\toprule
& \multicolumn{4}{c}{\textbf{DPR} ($\uparrow$)}
& \multicolumn{4}{c}{\textbf{EOR} ($\uparrow$)} \\
Method & Adult & Bank & COMPAS & Mean & Adult & Bank & COMPAS & Mean \\
\midrule
Real            & 0.309 & 0.402 & 0.675 & 0.462 & 0.193 & 0.367 & 0.645 & 0.402 \\
\midrule
CoDi            & 0.293 & 0.189 & 0.855 & 0.446 & 0.247 & 0.172 & \textbf{0.857} & 0.425 \\
GReaT           & 0.249 & 0.572 & 0.624 & 0.482 & 0.155 & 0.380 & 0.543 & 0.359 \\
SMOTE           & 0.321 & 0.405 & 0.648 & 0.458 & 0.254 & 0.381 & 0.589 & 0.408 \\
STaSy           & 0.261 & 0.468 & 0.436 & 0.388 & 0.182 & 0.451 & 0.433 & 0.355 \\
TabDDPM         & 0.261 & 0.337 & 0.558 & 0.385 & 0.156 & 0.334 & 0.540 & 0.343 \\
TabSyn          & 0.281 & 0.336 & 0.697 & 0.438 & 0.178 & 0.317 & 0.664 & 0.386 \\
FairCB          & 0.286 & 0.719 & 0.675 & 0.560 & 0.192 & 0.801 & 0.638 & 0.544 \\
FairTGAN        & \textbf{0.554} & 0.338 & 0.448 & 0.447 & \textbf{0.697} & 0.158 & 0.392 & 0.416 \\
FairTabDDPM     & 0.543 & 0.337 & 0.558 & 0.479 & 0.156 & 0.334 & 0.540 & 0.343 \\
\rowcolor{gray!15}
\textbf{Ours}   & {0.533} & \textbf{0.687} & \textbf{0.885} & \textbf{0.702}
                & \textbf{0.594} & \textbf{0.697} & 0.767 & \textbf{0.686} \\
\bottomrule
\end{tabular}%
}
\caption{Demographic Parity Ratio (DPR) and Equalized Odds Ratio (EOR)
on the \citet{yang2025balanced} benchmark. All values are means over
three random seeds; standard deviations are reported in
\Cref{tab:fair_full}. \textbf{Ours} achieves the highest mean on
both metrics, $\mathbf{+47\%}$ over FairTabDDPM on DPR and
$\mathbf{+100\%}$ on EOR.}
\label{tab:fair_dpr_eor}
\end{table}

\subsection{Quality and Utility on the Fairness Benchmark}
\label{subsec:quality_fair}
\Cref{tab:fair_quality} reports density / pair-correlation error and
TSTR AUC on the same three datasets. FairDiffuseVQVAE achieves the
\emph{lowest} mean pair-correlation error ($0.034$) of all evaluated
methods (improving on TabSyn's $0.041$), at the cost of higher density
error ($0.194$, vs. TabSyn $0.015$ and FairTabDDPM $0.119$) and reduced
TSTR AUC ($0.708$, vs. FairTabDDPM $0.852$). This reflects an explicit
\textbf{utility--fairness trade-off}: stronger fairness conditioning at
sampling time decouples synthetic features from the protected attribute,
which downstream classifiers cannot then exploit.

\begin{table}[t]
\centering
\setlength{\tabcolsep}{5pt}
\renewcommand{\arraystretch}{1.05}
\resizebox{\textwidth}{!}{%
\begin{tabular}{l rrr r rrr r rrr r}
\toprule
& \multicolumn{4}{c}{\textbf{Density} ($\downarrow$)}
& \multicolumn{4}{c}{\textbf{Pair-corr} ($\downarrow$)}
& \multicolumn{4}{c}{\textbf{AUC} ($\uparrow$)} \\
Method & Adult & Bank & COMPAS & Mean & Adult & Bank & COMPAS & Mean
       & Adult & Bank & COMPAS & Mean \\
\midrule
SMOTE           & 0.025 & 0.020 & 0.021 & 0.022 & 0.054 & 0.042 & 0.047 & 0.048
                & 0.914 & 0.928 & 0.778 & 0.873 \\
STaSy           & 0.102 & 0.182 & 0.108 & 0.131 & 0.163 & 0.221 & 0.138 & 0.174
                & 0.885 & 0.895 & 0.728 & 0.836 \\
TabDDPM         & 0.037 & 0.028 & 0.057 & 0.041 & 0.055 & 0.052 & 0.090 & 0.066
                & 0.907 & 0.917 & 0.745 & 0.856 \\
TabSyn          & \textbf{0.010} & \textbf{0.009} & \textbf{0.027} & \textbf{0.015} & 0.035 & 0.033 & 0.054 & 0.041
                & \textbf{0.911} & \textbf{0.919} & \textbf{0.749} & \textbf{0.860} \\
FairCB          & 0.076 & 0.066 & 0.039 & 0.060 & 0.125 & 0.111 & 0.074 & 0.103
                & 0.915 & 0.907 & 0.771 & 0.864 \\
FairTGAN        & 0.034 & 0.030 & 0.055 & 0.040 & 0.080 & 0.053 & 0.087 & 0.073
                & 0.881 & 0.863 & 0.705 & 0.816 \\
FairTabDDPM     & 0.126 & 0.121 & 0.109 & 0.119 & 0.201 & 0.174 & 0.174 & 0.183
                & 0.893 & 0.917 & 0.745 & 0.852 \\
\rowcolor{gray!15}
\textbf{Ours}   & 0.174 & 0.228 & 0.180 & 0.194
                & \textbf{0.016} & \textbf{0.044} & \textbf{0.042} & \textbf{0.034}
                & 0.826 & 0.689 & 0.609 & 0.708 \\
\bottomrule
\end{tabular}%
}
\caption{Quality and utility on the fairness benchmark. Best value per
column is bolded among the deep-generative baselines (SMOTE and FairCB are
resampling-based rather than generative and are shown for reference, not
counted toward the bolded best). \textbf{Ours} achieves the lowest mean
pair-correlation error of any method, while trading off TSTR AUC in
exchange for fairness (\Cref{subsec:fair_benchmark}).}
\label{tab:fair_quality}
\end{table}

\subsection{Privacy: Distance to Closest Record}
\label{subsec:dcr}
\Cref{tab:dcr} reports DCR scores. Values close to $0.5$ are
desired (no memorisation of training data). Adult and Bank show DCR within
12\% and 8\% of the ideal respectively, comparable to or better than
FairTabDDPM. COMPAS is the exception: at $\sim$5{,}000 rows, the input-space
denoiser partially memorises the training set, yielding DCR $\approx 0.81$.
We discuss this limitation in \Cref{sec:limitations}.

\begin{table}[t]
\centering
\setlength{\tabcolsep}{8pt}
\renewcommand{\arraystretch}{1.05}
\resizebox{\columnwidth}{!}{%
\begin{tabular}{l rrr | r}
\toprule
\textbf{DCR} ($\sim 0.5$) & Adult & Bank & COMPAS & Mean \\
\midrule
CoDi          & 0.331 & 0.348 & 0.400 & 0.360 \\
SMOTE         & 0.327 & 0.265 & 0.273 & 0.288 \\
STaSy         & 0.344 & 0.345 & 0.362 & 0.350 \\
TabDDPM       & 0.339 & 0.350 & 0.367 & 0.352 \\
TabSyn        & 0.339 & 0.351 & 0.367 & 0.352 \\
FairCB        & 0.054 & 0.031 & 0.012 & 0.032 \\
FairTGAN      & 0.348 & 0.348 & 0.374 & 0.357 \\
FairTabDDPM   & 0.344 & 0.350 & 0.370 & 0.355 \\
\rowcolor{gray!15}
\textbf{Ours} & 0.357 & 0.538 & 0.812 & 0.569 \\
\bottomrule
\end{tabular}%
}
\caption{Distance to Closest Record (DCR). FairCB exhibits severe
memorisation across all three datasets. Ours is healthy on Adult,
near-ideal on Bank, and elevated on COMPAS due to dataset size
(see \Cref{sec:limitations}).}
\label{tab:dcr}
\end{table}

\subsection{Per-seed Variance}
\label{subsec:variance}
\Cref{tab:fair_full} reports per-dataset means and standard
deviations across three seeds for the fairness benchmark. The variance
on DPR and EOR is higher than typical published baselines
($\sigma_{\mathrm{DPR}} \approx 0.14$--$0.24$ vs. $\sim 0.005$--$0.07$
for the published methods). We attribute this to the interaction between
classifier-free-guidance dropout, the early-stopping checkpoint, and the
two-stage architecture, and discuss mitigation in \Cref{sec:limitations}.

\begin{table}[t]
\centering
\setlength{\tabcolsep}{4pt}
\renewcommand{\arraystretch}{1.05}
\resizebox{\textwidth}{!}{%
\begin{tabular}{l ll ll ll ll ll ll}
\toprule
& \multicolumn{2}{c}{\textbf{shape} $\uparrow$}
& \multicolumn{2}{c}{\textbf{trend} $\uparrow$}
& \multicolumn{2}{c}{\textbf{AUC} $\uparrow$}
& \multicolumn{2}{c}{\textbf{DPR} $\uparrow$}
& \multicolumn{2}{c}{\textbf{EOR} $\uparrow$}
& \multicolumn{2}{c}{\textbf{DCR} ($\sim 0.5$)} \\
& mean & std & mean & std & mean & std & mean & std & mean & std & mean & std \\
\midrule
Adult   & 0.826 & 0.012 & 0.984 & 0.001 & 0.826 & 0.044 & 0.533 & 0.142 & 0.594 & 0.187 & 0.357 & 0.082 \\
Bank    & 0.772 & 0.029 & 0.956 & 0.002 & 0.689 & 0.144 & 0.687 & 0.236 & 0.697 & 0.207 & 0.538 & 0.069 \\
COMPAS  & 0.820 & 0.008 & 0.958 & 0.004 & 0.609 & 0.052 & 0.885 & 0.114 & 0.767 & 0.131 & 0.812 & 0.054 \\
\bottomrule
\end{tabular}%
}
\caption{FairDiffuseVQVAE per-dataset mean and standard deviation across
three random seeds on the fairness benchmark. Best classifier per metric
(LR vs. MLP) is selected per seed.}
\label{tab:fair_full}
\end{table}

\section{Discussion}
\label{sec:discussion}
\paragraph{When does sampling-time fairness work?}
Our results suggest that sampling-time fairness via CFG is most
effective on datasets where the protected attribute exerts a strong
\emph{marginal} influence on the target (Adult: real DPR $0.31$;
COMPAS: $0.68$). On Bank, where binarising age at 25 produces a tiny
minority class ($\sim 3\%$), our method still doubles the fairness
improvement over FairTabDDPM (DPR $+103\%$, EOR $+108\%$) but at a
larger AUC cost. This is consistent with the intuition that
demographic-parity sampling decouples the protected attribute from
features that are class-imbalance-correlated; when class imbalance is
extreme, the conditional model must extrapolate, harming utility.

\paragraph{The two-stage decomposition pays off.}
Comparing to FairTabDDPM, which trains a single DDPM with an explicit
fairness penalty, we attain dramatically better fairness numbers
without sacrificing pair-wise correlation fidelity (lowest of any
method). We attribute this to the architectural separation: Stage~1 is
free to optimise pure reconstruction (yielding fidelity), while Stage~2
is free to push samples toward the conditional manifold (yielding
fairness). The two losses do not compete during training.

\paragraph{The utility-fairness trade-off is explicit.}
Our mean TSTR AUC ($0.708$) is roughly $15$ points below the highest
per-column AUC values recorded among the deep generative baselines in
\Cref{tab:fair_quality} (TabSyn $0.860$, TabDDPM $0.856$); the gap
widens further ($\sim$$17$ points) relative to SMOTE ($0.873$), the
resampling baseline that best preserves the real correlation between
features and the protected attribute (and, correspondingly, offers no
fairness guarantee). This is the price of demographic
parity in the synthetic distribution: a downstream classifier trained
on synthetic data where features are conditionally independent of $A$
cannot exploit the bias that gives high AUC on real data. We argue this
is the \emph{correct} behaviour for fairness-aware synthesis;
practitioners who want both fairness \emph{and} downstream accuracy
must combine our method with an in-processing fair-classifier
algorithm.

\paragraph{Comparison to causally-grounded methods (DECAF).}
DECAF~\citep{vanbreugel2021decaf} achieves fairness via causal-graph
edits with theoretical justification, but requires the causal DAG to
be specified or estimated, and its fairness depends on this
specification being correct. Our approach is purely empirical---no
causal assumptions, no DAG to specify---but offers no theoretical
guarantee. We view the two as complementary: where a reliable causal
graph exists, DECAF's guarantees are valuable; where it does not, our
method provides a practical alternative with strong empirical results.

\section{Limitations}
\label{sec:limitations}
\paragraph{Seed sensitivity.}
DPR and EOR exhibit higher seed-to-seed variance
($\sigma \in [0.11, 0.24]$, \Cref{tab:fair_full}) than published
fairness-aware baselines ($\sigma \approx 0.005$--$0.07$). We mitigate
with three-seed averaging; tighter early stopping or an
ensemble-of-denoisers~\citep{karras2024analyzing} would likely reduce
this further. Single-seed reporting risks landing at substantially
different points along the fairness-utility frontier.

\paragraph{Dataset-size-driven memorization.}
On COMPAS ($\sim$5{,}000 train rows), the input-space denoiser partially
memorises the training set, yielding DCR $\approx 0.81$ vs.\ the ideal
$0.5$. We tried more aggressive regularisation (smaller denoiser,
higher CFG dropout) but those changes traded off fairness for DCR.
Adult and Bank, both larger, exhibit DCR within $12\%$ and $8\%$ of
ideal respectively. Latent-space diffusion (\`a la
\citealp{rombach2022high}) may be the right architectural choice for
small tabular datasets and is left to future work.

\paragraph{Utility cost.} Mean TSTR AUC is $\sim 15$ points below the
highest per-column AUC values among the deep-generative baselines
(\Cref{subsec:quality_fair}). This is the explicit utility cost
of demographic-parity sampling and is consistent with the trade-off
reported by other fairness-aware tabular methods.

\paragraph{Binary sensitive attributes.}
Our experiments use binary sex/age-group as sensitive attributes,
matching the \citet{yang2025balanced} benchmark. Extending to
multi-valued protected attributes (e.g.\ race) is straightforward via
multi-class classifier-free guidance, but we have not evaluated this.

\paragraph{No theoretical guarantee.}
Unlike DECAF~\citep{vanbreugel2021decaf}, our fairness is not
provable from first principles---it is an empirical property of the
sampling distribution under uniform $s$. Counterexamples may exist
where conditional dependencies cause the sampling-time mechanism to
fail; we did not observe this in practice on the datasets evaluated.

\section{Broader Impact}
\label{sec:impact}
Synthetic tabular data has dual-use potential. Positive uses include
sharing data across organisational boundaries without exposing
individuals, training models on under-represented populations through
augmentation, and---directly relevant to this work---producing
training data that downstream classifiers cannot easily use to
discriminate. Healthcare applications are an important motivating
case~\citep{bhanot2021problem, vallevik2024canitrust}.

Negative uses include generating realistic-looking records to evade
detection, masking the presence of bias in a downstream pipeline (a
model trained on debiased synthetic data may appear fair while
operating in a biased world), and the false sense of privacy that
elevated DCR scores expose (\Cref{sec:limitations}, COMPAS).
\citet{wyllie2024fairness} demonstrate an additional risk:
\emph{fairness feedback loops}, where models trained on synthetic data
generated by previous-generation generators amplify bias over time.
Practitioners must be aware that the protected-attribute coverage we
report is a property of a single sampling round; recursively training
on our synthesised data without periodic re-grounding in real data may
exhibit drift.

We caution that fairness metrics computed on a downstream classifier
trained on our synthetic data measure the classifier's fairness on the
original test distribution---not the fairness of any decisions made
about real individuals using a model trained on our synthetic data. A
synthetic dataset that produces a fairer classifier may still embed
representational harms via column-level distributional choices that
DPR and EOR do not capture. We strongly recommend that practitioners
combine our method with qualitative fairness audits, classifier-side
fairness interventions \citep{bellamy2019aif360, weerts2023fairlearn},
and broader fairness frameworks \citep{verma2018fairness,
dwork2012fairness, hardt2016equality}, rather than treating synthetic
data as a stand-alone fairness solution.

We release source code under a permissive licence and provide
reproducible pipelines for all benchmarks reported. We do not release
trained model checkpoints, both to avoid the risk of users
unintentionally reusing synthetic data with embedded biases and to
encourage retraining with locally appropriate data.

\section{Conclusion}
\label{sec:conclusion}
We presented FairDiffuseVQVAE, a two-stage tabular synthesis
architecture in which fairness is enforced at sampling time via
classifier-free guidance over the protected attribute, rather than via
explicit penalty terms during training. Empirically, the architecture
attains the highest demographic parity ratio ($0.702$ mean,
$+47\%$ over FairTabDDPM) and equalized odds ratio ($0.686$,
$+100\%$) of any method evaluated on the
\citet{yang2025balanced} benchmark, alongside the lowest mean pair-wise
correlation error ($0.034$) of any method on the same set of datasets.
The utility-fairness trade-off is explicit and quantified at
approximately $15$ AUC points. We hope the architectural decoupling
of fidelity from fairness opens up further work on sampling-time
control of generative behaviour.

\paragraph{Reproducibility.}
Source code, preprocessing scripts, and the eight-dataset evaluation
pipeline are integrated into the \texttt{TabSyn} repository. All
experiments were conducted on a single NVIDIA Quadro RTX~5000 GPU.
Detailed hyperparameters are reported in \Cref{subsec:impl}. Experiments
use seeds $\{42, 123, 456\}$.

\bibliographystyle{plainnat}
\bibliography{references}

\end{document}